\documentclass[journal]{IEEEtran}
\usepackage{cite}
\usepackage{amsmath,amssymb,amsfonts}
\usepackage{subfigure}
\usepackage{algorithmic}
\usepackage{graphicx}
\usepackage{textcomp}
\usepackage{xcolor}
\usepackage{threeparttable}
\usepackage{multirow}
\usepackage{algorithm}
\usepackage{algorithmic}
\usepackage{booktabs,tabulary}
\usepackage{enumitem}
\usepackage[bookmarks=true]{hyperref}
\usepackage[hyphenbreaks]{breakurl}
\ifCLASSINFOpdf
\else
\fi
\begin{document}
%
\title{Mobile Robot Path Planning in Dynamic Environments through Globally Guided Reinforcement Learning}
%
%
%

\author{Binyu~Wang, Zhe~Liu, Qingbiao~Li,
        Amanda~Prorok,~\IEEEmembership{Member,~IEEE}
\thanks{This work was supported by the Engineering and Physical Sciences Research Council (grant EP/S015493/1), and ARL DCIST CRA W911NF- 17-2-0181.}
\thanks{B.~Wang is with the Department of Mechanical and Automation Engineering, The Chinese University of Hong Kong, Hong Kong (e-mail: zbwby819@sina.com). Z.~Liu, Q.~Li and A.~Prorok are with the Department of Computer Science and Technology, University of Cambridge, Cambridge CB3 0FD, U.K. (e-mail: \{zl457, ql295, asp45\}@cam.ac.uk). \emph{Corresponding author: Z.~Liu.}}
}

\maketitle

\begin{abstract}
Path planning for mobile robots in large dynamic environments is a challenging problem, as the robots are required to efficiently reach their given goals while simultaneously avoiding potential conflicts with other robots or dynamic objects.
In the presence of dynamic obstacles, traditional solutions usually employ re-planning strategies, which re-call a planning algorithm to search for an alternative path whenever the robot encounters a conflict. However, such re-planning strategies often cause unnecessary detours.
To address this issue, we propose a learning-based technique that exploits environmental spatio-temporal information. Different from existing learning-based methods, we introduce a \textit{globally guided reinforcement learning} approach (\textbf{G2RL}), which incorporates a novel reward structure that generalizes to arbitrary environments.
We apply G2RL to solve the multi-robot path planning problem in a fully distributed reactive manner.
We evaluate our method across different map types, obstacle densities, and the number of robots. Experimental results show that G2RL generalizes well, outperforming existing distributed methods, and performing very similarly to fully centralized state-of-the-art benchmarks.
\end{abstract}


%
\IEEEpeerreviewmaketitle

\section{Introduction}
\IEEEPARstart{P}{ath} planning is one of the fundamental problems in robotics. It can be formulated as: given a robot and a description of the environment, plan a conflict-free path between the specified start and goal locations. Traditionally, there are two different versions: off-line planning, which assumes static obstacles and perfectly known environments, and on-line planning, which focuses on dealing with dynamic obstacles and partially known environments \cite{smierzchalski2005path}. Traditional off-line planning algorithms \cite{cohen2018fastmap} cannot be directly utilized for solving on-line path planning tasks as they assume that the obstacles are static. One strategy is to plan an initial path and invoke re-planning whenever its execution becomes infeasible \cite{stentz1997optimal}. However, re-planning will suffer from time inefficiency (frequent re-planning) and path inefficiency (oscillating movements and detours) due to the absence of motion information of dynamic obstacles. Furthermore, re-planning strategies may fail in the presence of robot deadlocks. Instead of re-planning, some methods include an extra time dimension to avoid potential conflicts \cite{liu2020}. However, this approach increases the number of states to be searched, and additionally requires the knowledge of the future trajectories of dynamic obstacles. If the future movements of obstacles are unknown, one may attempt to model the behavior of dynamic obstacles and predict their paths \cite{phillips2011sipp}. Yet, separating the navigation problem into disjoint prediction and planning steps can lead to the `freezing robot' problem. In that case, the robot will fail to find any feasible action as the predicted paths could mark a large portion of the space as untraversable.

Recently, learning-based approaches have been studied to address on-line planning in dynamic environments \cite{sartoretti2019primal,Qingbiao2019}. This is popularized by the seminal work \cite{mnih2015human}, which utilized deep neural networks for the function estimation of value-based reinforcement learning (RL). Although RL has demonstrated outstanding performance in many applications, several challenges still impede its contribution to the path planning problem. First of all, when the environment is extremely large, the reward becomes sparse, inducing an increased training effort and making the overall learning process inefficient~\cite{Sutton1998Reinforcement}. Another challenge is the over-fitting issue. The robot is often limited to training environments and shows poor generalizability to unseen environments~\cite{pan2019you}. Most recent approaches still show difficulties in scaling to arbitrarily large multi-robot systems~\cite{Qingbiao2019}, as the sizes of the robot state, joint action, and joint observation spaces grow exponentially with the number of robots~\cite{Sun2020}. Thus, the efficiency, generalizability, and scalability of existing RL-based planners can still not fulfill the requirements of many applications.

In order to overcome the above challenges, we develop a hierarchical path-planning algorithm that combines a {\it global guidance} and a {\it local RL-based planner}. Concretely, we first utilize a global path planning algorithm (for example, A*) to obtain a globally optimal path, which we refer to as the {\it global guidance}. During robot motion, the {\it local RL-based planner} generates robot actions by exploiting surrounding environmental information to avoid conflicts with static and dynamic obstacles, while simultaneously attempting to follow the fixed global guidance.
Our main contributions include:
\begin{itemize}[leftmargin=*]
\item We present a hierarchical framework that combines global guidance and local RL-based planning to enable end-to-end learning in dynamic environments.
The local RL planner exploits both spatial and temporal information within a local area (e.g., a field of view) to avoid potential collisions and unnecessary detours. Introducing global guidance allows the robot to learn to navigate towards its destination through a \textit{fixed-sized} learning model, even in large-scale environments, thus ensuring scalability of the approach.

\item We present a novel reward structure that provides dense rewards, while not requiring the robot to strictly follow the global guidance at every step, thus encouraging the robot to explore all potential solutions. In addition, our reward function is independent of the environment, thus enabling scalability as well as generalizability across environments.

\item We provide an application of our approach to multi-robot path planning, whereby robot control is fully distributed and can be scaled to an arbitrary number of robots.

\item Experimental results show that our single-robot path planning approach outperforms local and global re-planning methods, and that it maintains consistent performance across numerous scenarios, which vary in map types and number of obstacles. In particular, we show that our application to multi-robot path planning outperforms current state-of-the-art distributed planners. Notably, the performance of our approach is shown to be comparable to that of centralized approaches, which, in contrast to our approach, assume global knowledge (i.e., trajectories of all dynamic objects).
\end{itemize}

\section{Background and Related Work}

{\it{Traditional path planning approaches.}}
Path planning can be divided into two categories: global path planning and local path planning \cite{Peng2011A}. The former approach includes graph-based approaches (for example, Dijkstra and A* \cite{dijkstra1959note}) and sampling-based approaches (for example, RRT and its variant \cite{Brian2018}), in which all the environmental information is known to the robot before it moves. For local path planning, at least a part or almost all the information on the environment is unknown. Compared to global path planners, local navigation methods can be very effective in dynamic environments. However, since they are essentially based on the fastest descent optimization, they can easily get trapped in a local minimum \cite{Wang2002Hybrid}. A promising solution is to combine the local planning with global planning, where the local path planner is responsible for amending or optimizing the trajectory proposed by the global planner. For instance, \cite{Brock1999High} proposed a global dynamic window approach that combines path planning and real-time obstacle avoidance, allowing robots to perform high-velocity, goal-directed, and reactive motion in unknown and dynamic environments. Yet their approach can result in highly sub-optimal paths. The authors in \cite{Mehta2016Autonomous} adopt multi-policy decision making to realize autonomous navigation in dynamic social environments. However, in their work, the robot’s trajectory was selected from closed-loop behaviors whose utility can be predicted rather than explicitly planned.

{\it{Learning based approaches.}}
Benefiting from recent advances in deep learning techniques, learning-based approaches have been considered as a promising direction to address path planning tasks. Reinforcement Learning (RL) has been implemented to solve the path planning problem successfully, where the robot learns to complete the task by trial-and-error. Traditionally, the robot receives the reward after it reaches the target location \cite{Sutton1998Reinforcement}.
As the environment grows, however, the robot needs to explore more states to receive rewards.
Consequently, interactions become more complex, and the learning process becomes more difficult. Other approaches apply Imitation Learning (IL) to provide the robot dense rewards to relieve this issue \cite{sartoretti2019primal, Riviere2020, Qingbiao2019}. However, basing the learning procedure on potentially biased expert data may lead to sub-optimal solutions~\cite{silver2017mastering, bhardwaj2017learning}. Compounding this issue, the robot only receives rewards by strictly following the behavior of expert demonstrations, limiting exploration to other potential solutions. Also, over-fitting remains a problem. This is clearly exemplified in~\cite{pan2019you}, where the robot follows the previously learned path, even when all obstacles have been removed from the environment.

\begin{figure*}[t]
\centering
\includegraphics[width=1\textwidth]{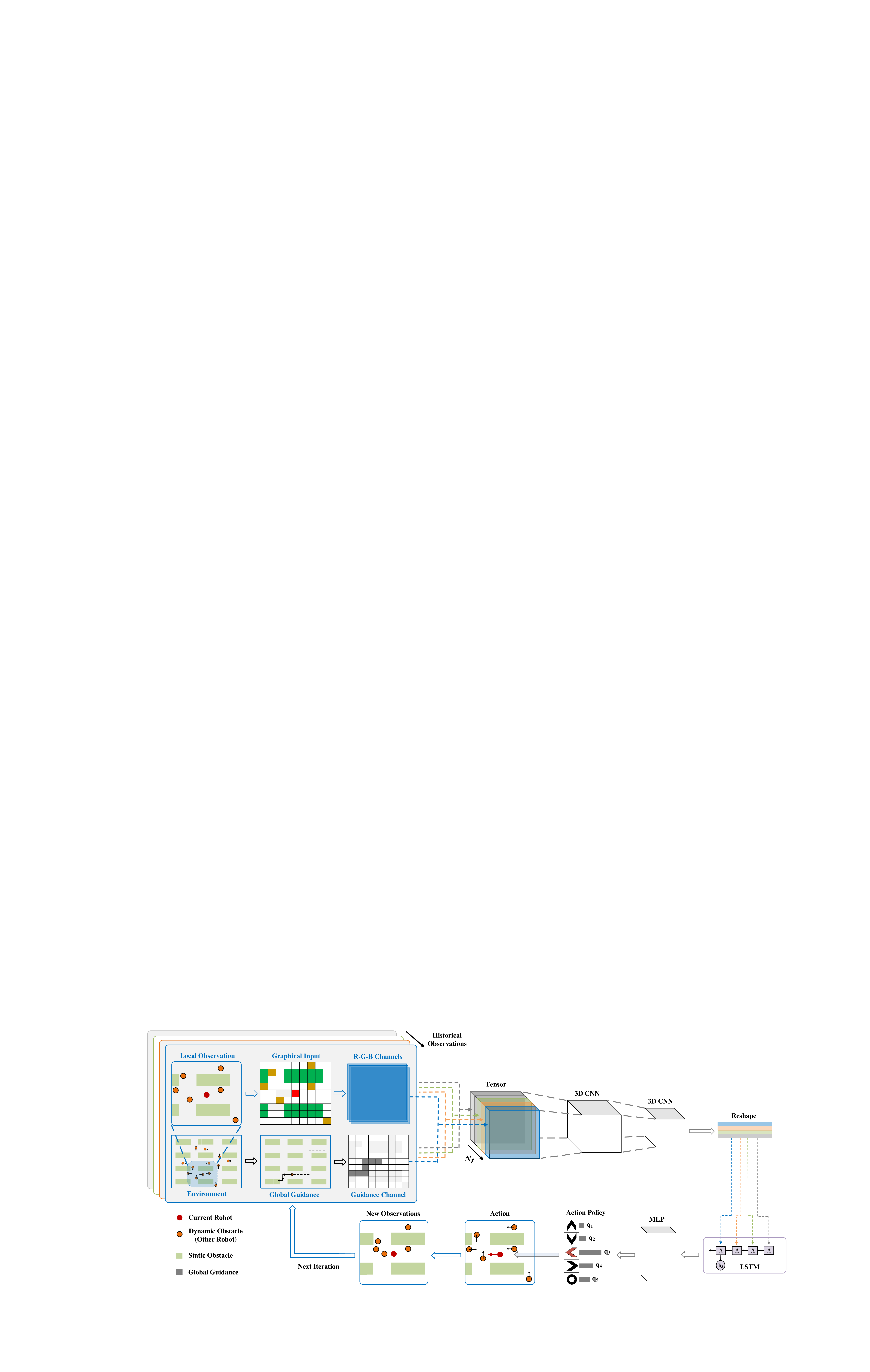}
\caption{The overall structure of G2RL. The input of each step is the concatenation of the transformed local observation and the global guidance information. A sequence of historical inputs is combined to build the input tensor of the deep neural network, which outputs a proper action for the robot.}
\label{Fig_overallsystem}
\vspace{-0.4cm}
\end{figure*}

\section{Problem Description}

{\it Environment representation}. Consider a 2-dimensional discrete environment $\mathcal{W}\subseteq\mathbb{R}^2$ with size $H\times W$ and a set of $N_s$ static obstacles $\mathcal{C}_s = \{s_{1}, ..., s_{N_s}\}$, where $s_i\subset\mathcal{W}$ denotes the $i^{th}$ static obstacle.
The free space $\mathcal{W}\setminus\mathcal{C}_s$ is represented by a roadmap ${G}=\langle\mathcal{C}_f,\mathcal{E}\rangle$, where $\mathcal{C}_f=\{c_1,...,c_{N_f}\}=\mathcal{W}\setminus\mathcal{C}_s$ represents the set of free cells and $e_{ij}=(c_i,c_j)\subset\mathcal{E}$ represents the traversable space between free cells $c_i$ and $c_j$ that does not cross any other cell (the minimum road segment).
The set of dynamic obstacles $\mathcal{C}_d(t) = \{d_{1}(t), ..., d_{N_d}(t)\}$ denotes the position of $N_d$ dynamic obstacles at time $t$, where $\forall i,j,t$, $d_i(t)\subset\mathcal{C}_f$, $(d_i(t),d_i(t+1))\subset\mathcal{E}$ or $d_i(t)=d_i(t+1)$, and $d_i(t)\neq d_j(t)$. In addition, if $d_i(t+1)=d_j(t)$, then $d_j(t+1)\neq d_i(t)$, i.e., any motion conflict should be avoided.

{\it Global guidance}. A {\it traversable path} $\mathcal{G}=\{g(t)\}$ is defined by the following rules: 1) the initial location $g(0)=c_{start}$ and there is a time step $t_f$ that $\forall t\geq t_f$, $g(t)=c_{goal}$. Note that, $c_{start}, c_{goal} \in \mathcal{C}_f$; 2) $\forall t$, $g(t)\in \mathcal{C}_f$, 
$(g(t),g(t+1))\in \mathcal{E}$. The {\it global guidance} $\mathcal{G}^*$ is the shortest traversable path, defined as $\mathcal{G}^*=\arg\min_{t_f}{\mathcal{G}}$, between the initial location $c_{start}$ and the goal $c_{goal}$. Note that the global guidance may not be unique---in this paper, we randomly choose one instance.

{\it Assumptions}. We assume that the robot knows the information of all the static obstacles and calculates the global guidance at the start of each run. Note that the global guidance is only calculated once and remains the same. During robot motion, we assume that the robot can obtain its global location in the environment and acquire global guidance information. However, we do not assume that the trajectories of dynamic obstacles are known to the robot. The robot can only obtain the current location of dynamic obstacles when they are within its local field of view.

{\it Local observation}. The robot has a local field of view (FOV) within which it observes the environment. More specifically, at each time step $t$, the robot collects the local observation $O_t=\{o^f_t,o^s_t,o^d_t\}$ which is a collection of the location of free cells $o^f_t$, static obstacles $o^s_t$ and dynamic obstacles $o^d_t$, within the local FOV with the size of $H_l \times W_l$. In addition, we also define a local segment of the global guidance $\mathcal{G}^*$ as the local path segment $\mathcal{G}^*_t$, which is located within the robot's local FOV. Both $O_t$ and $\mathcal{G}^*_t$ are considered as input information at each time step $t$.

{\it Robot action}. The robot action set is defined as $\mathcal{A}=\{\mathrm{Up, Down, Left, Right, Idle}\}$, i.e., at each time step, the robot can only move to its neighboring locations or remain in its current location.

{\it Objective}. Given as input the local segment $\mathcal{G}^*_t$ of the global guidance, the current local observation $\mathcal{O}_t$, and a history of local observations $\mathcal{O}_{t-1}$,..., $\mathcal{O}_{t-(N_t-1)}$, output an action $a_t\subset \mathcal{A}$ at each time step $t$ that enables the robot to move from the start cell $c_{start}$ to the goal cell $c_{goal}$ with the minimum number of steps while avoiding conflicts with static and dynamic obstacles.

\section{RL-Enhanced Hierarchical Path Planning}
In this section, we first describe the overall system structure and then present details of our approach.
\subsection{System Structure}
Figure \ref{Fig_overallsystem} illustrates the overall system structure, which shows that our local RL planner contains four main modules:

\begin{enumerate}[leftmargin=*]
\item {\it Composition of network input:} Firstly, we transform the local observation $O_t$ into a graphic and use its three channels (RGB data) in combination with a guidance channel to compose the current input. Then a sequence of historical inputs is used to build the final input tensor;
\item {\it Spatial processing:} Secondly, we utilize a series of 3D CNN layers to extract features from the input tensor, then reshape the feature into $N_t$ one-dimensional vectors;
\item {\it Temporal processing:} Thirdly, we use an LSTM layer to further extract the temporal information by aggregating the $N_t$ vectors;
\item {\it Action policy:} Finally, we use two fully connected (FC) layers to estimate the quality $q_i$ of each state-action pair and choose the action $a_i\subset \mathcal{A}$ with $\max_i q_i$.
\end{enumerate}

\subsection{Global Guidance and Reward Function}

The main role of global guidance is to provide long-term global information. By introducing this information, the robot receives frequent feedback signals in arbitrary scenarios, no matter how large the environment and how complex the local environments are. We achieve this by proposing a novel reward function that provides dense rewards while simultaneously not requiring the robot to follow the global guidance strictly. In this manner, we encourage the robot to explore all the potential solutions while also promoting convergence of the learning-based navigation.

More specifically, at each step, the reward function offers: 1) A small negative reward $r_1<0$ when the robot reaches a free cell which is not located on the global guidance; 2) A large negative reward $r_1+r_2$ when the robot conflicts with a static obstacle or a dynamic obstacle, where $r_2<r_1<0$; 3) A large positive reward $r_1+N_e\times r_3$ when the robot reaches one of the cells on the global guidance path, where $r_3>|r_1|>0$ and $N_e$ is the number of cells removed from the global guidance path, between the point where the robot first left that path, to the point where it rejoins it. The reward function can be defined formally as:
\begin{equation}\label{Eq_reward}
R(t) = \left\{
\begin{aligned}
& r_1                  &  \mathrm{if} \; c_r(t+1)\in\mathcal{C}_f\setminus\mathcal{G}^* \\
& r_1 + r_2             &  \mathrm{if} \; c_r(t+1)\in\mathcal{C}_s\cup\mathcal{C}_d(t+1) \\
& r_1 + N_e\times r_3    &  \mathrm{if} \; c_r(t+1) \in\mathcal{G}^*\setminus\mathcal{C}_d(t+1)
\end{aligned}
\right.
\end{equation}
where $c_r(t+1)$ is the robot location after the robot takes the action $a_t$ at time $t$,  $R(t)$ is the reward value of action $a_t$.

\begin{figure*}[t]
\centering
\includegraphics[width=0.8\textwidth]{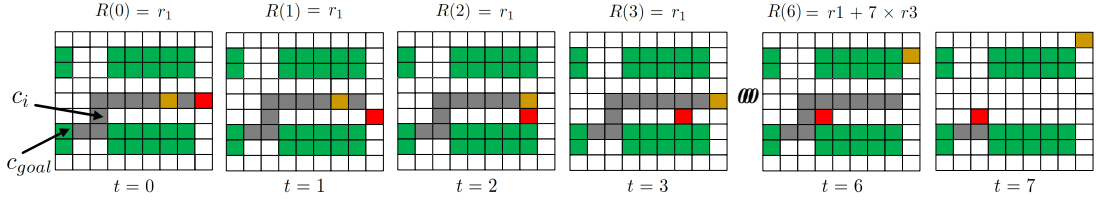}
\caption{An illustration of our reward function. The green, black, yellow and red cells represent the static obstacle, the global guidance, the dynamic obstacle and the robot, respectively. At $t=7$, the robot reaches a global guidance cell $c_i$ and receives the reward based on the number of eliminated guidance cells.}
\label{Fig_reward}
\vspace{-0.4cm}
\end{figure*}

Figure \ref{Fig_reward} shows an example of our reward function. At $t=0$, since there is a dynamic obstacle in front, our RL planner moves the robot to the lower cell to avoid conflict. From $t=1$ to $t=6$, the robot does not need to return to the global guidance path immediately, but can continue to move left until it reaches one cell $c_i$ located on the global guidance path at $t=7$. Here $R(0)=R(1)=R(2)=R(3)=R(4)=R(5)=r_1$ and $R(6)=r_1+7\times r_3$, since at $t=7$, $N_e=7$ cells have been removed in the global guidance. We remove the path segment up to $c_i$ as soon as the robot obtains the reward $R(6)$ to ensure that the reward of each cell on the global guidance path can only be collected once. The removed guidance cells will be marked as normal free cells (white cells) in the graphic images inputted in the following iterations. In contrast to IL-based methods, we do not require the robot to strictly follow the global guidance at each step, since the robot receives the same cumulative reward as long as it reaches the same guidance cell given from the same start cell. As a result, our model can also be trained from scratch without IL, which circumvents potentially biased solutions~\cite{silver2017mastering}. In the training process, we stop the current episode once the robot deviates from the global guidance too much, namely, if no global guidance cell can be found in the FOV of the robot.

\subsection{Local RL Planner}

As shown in Figure \ref{Fig_overallsystem}, we transform a robot's local observation into an {\it observation image} defined as: 1) The center pixel of the image corresponds to the current robot position, the image size is the same as the robot's FOV $H_l\times W_l$, i.e., one pixel in the image corresponds to one cell in the local environment; 2) All the static and dynamic obstacles observed are marked in the image, where we use one color to represent static obstacles and another color to denote dynamic ones. In addition, a guidance channel that contains the local path segment $\mathcal{G}^*_t$ of the global guidance is introduced to combine with the three channels (RGB data) of the observation image, to compose the current input $\mathcal{I}_t$. Our RL planner takes the sequence $\mathcal{S}_t=\{\mathcal{I}_t,\mathcal{I}_{t-1}...\mathcal{I}_{t-(N_t-1)}\}$ as the inputs at step $t$.

We use Double Deep Q-Learning (DDQN) \cite{VanDeep} for our RL planner. At time step $t$, the target value $Y_t=R(t)$, if $c_{t+1}= c_{goal}$, otherwise,
\begin{equation}\label{Eq_targetvalue}
Y_t = R(t)+\gamma Q(\mathcal{S}_{t+1}, \arg\max_{a_{t}} Q(\mathcal{S}_{t+1}, a_{t};\theta );\theta ^{-})
\end{equation}
where $Q(.)$ is the quality function, $\theta $ and $\theta ^{-}$ are the current and target network parameters respectively, $\gamma$ is the discount value and $R(t)$ is defined in \eqref{Eq_reward}.
To update the parameters $\theta$, we sample $N_b$ transitions from the replay buffer, and define the loss $\mathcal{L}$ as:
\begin{equation}
\mathcal{L} = \frac{1}{N_b}\sum_{j=1}^{N_b}(Y_t^j - Q(\mathcal{S}_t^j, a_{t}^j; \theta ))^2
\end{equation}
As shown in Figure \ref{Fig_overallsystem}, our model is comprised of 3D CNN layers followed by LSTM and FC layers.
The input is a five-dimensional tensor with size $N_t \times H_l \times W_l \times 4$, where $N_b$ is the batch size sampled from the replay buffer. We first use 3D CNN layers to extract the spatial information. The size of the convolutional kernel is $1 \times 3 \times 3$. We stack one CNN layer with kernel stride $1 \times 1 \times1$ and another with stride $1 \times 2 \times 2$ as a convolutional block, which repeats $N_c$ times for downsampling in the spatial dimension.
The embeddings extracted by the CNN layers are reshaped into $N_t$ one-dimensional vectors and fed into the LSTM layer to extract temporal information. Two FC layers are attached to the LSTM layer, where the first layer (followed by a \texttt{ReLU} activation function) reasons about the extracted information and the second layer directly outputs the value $q_i$ of each state-action pair $(\mathcal{S}_t, a_i)$ with a \texttt{Linear} activation function.

\subsection{Application to Reactive Multi-Robot Path Planning}
The main idea of our local RL planner is to encourage the robot to learn how to avoid surrounding obstacles while following global guidance. Since robots only consider local observations and global guidance, and do not need to explicitly know any trajectory information nor motion intentions of other dynamic obstacles or robots, the resulting policy can be copied onto any number of robots and, hence, scales to arbitrary numbers of robots.
Based on the aforementioned rationale, our approach is easily extended to resolving the multi-robot path planning problem in a fully distributed manner, whereby dynamic obstacles are modeled as independent mobile robots. Each robot considers its own global guidance and local observations to generate actions. Since we do not require communication among robots, this is equivalent to an uncoordinated reactive approach. Note that the above extension is fully distributed, can be trained for a single agent (i.e., robot) and directly used by any number of other agents.

\section{Implementation}
In this section, we introduce the network parameters and describe our training and testing strategies.

\subsection{Model Parameters}
In the experiments, we use the A* algorithm to generate the global guidance. The default parameters are set as follows: robot local FOV size $H_l=W_l=15$, the length of input sequence $N_t=4$, the reward parameters $r_1=-0.01$, $r_2=-0.1$, and $r_3=0.1$. The convolutional block is repeated $N_c=3$ times with input batch size $N_b=32$. The activation functions are \texttt{ReLU}. We use 32 convolution kernels in the first convolutional block and double the number of kernels after each block. After the CNN layers, the shape of feature maps is $4 \times 2 \times 2 \times 128$. In the LSTM layer and the two FC layers, we use 512, 512, and 5 units, respectively.

\begin{figure}[!t]
	\centering
    \subfigure[Regular map]{\includegraphics[width=0.3\columnwidth]{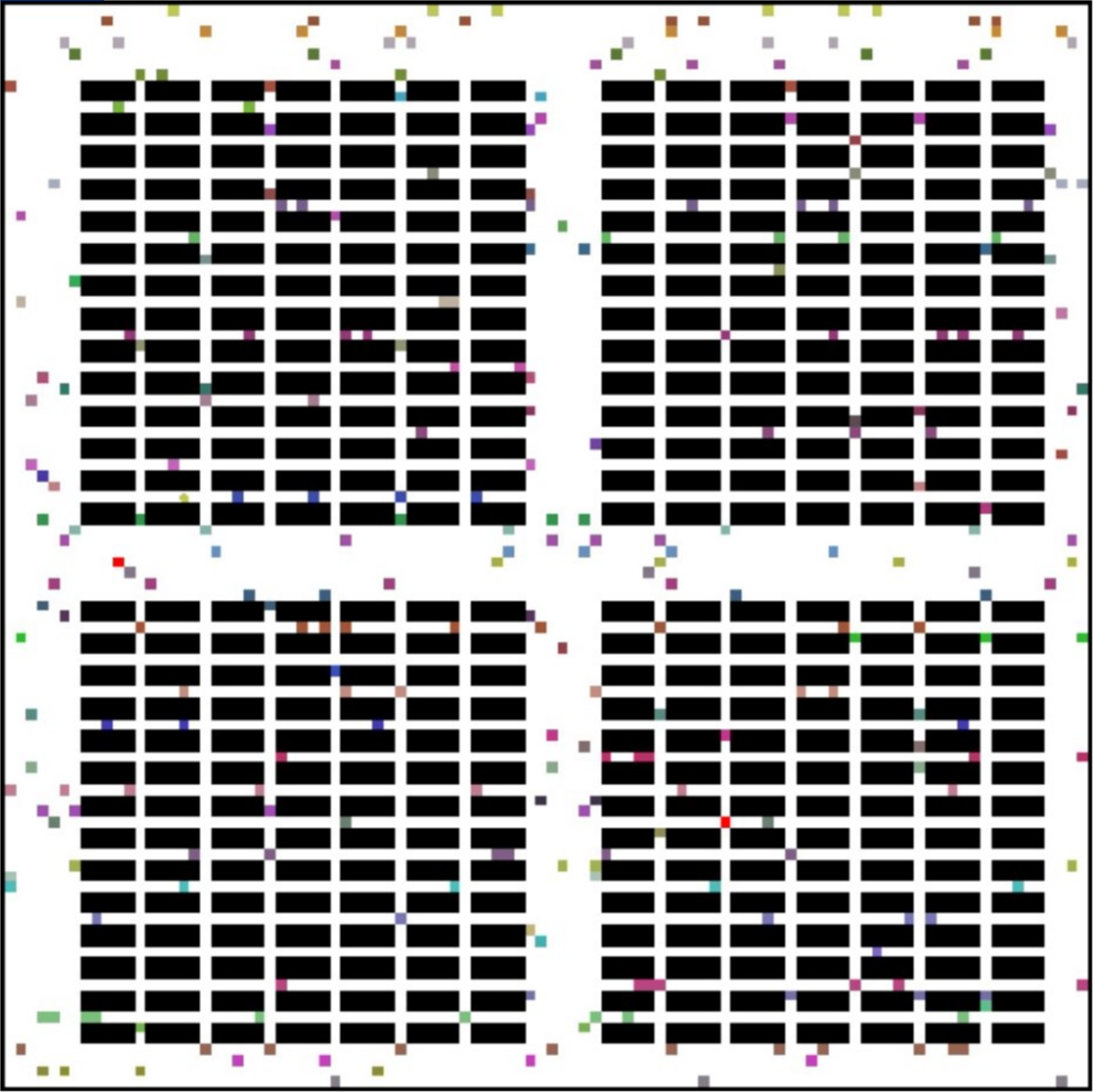}}
    \subfigure[Random map]{\includegraphics[width=0.3\columnwidth]{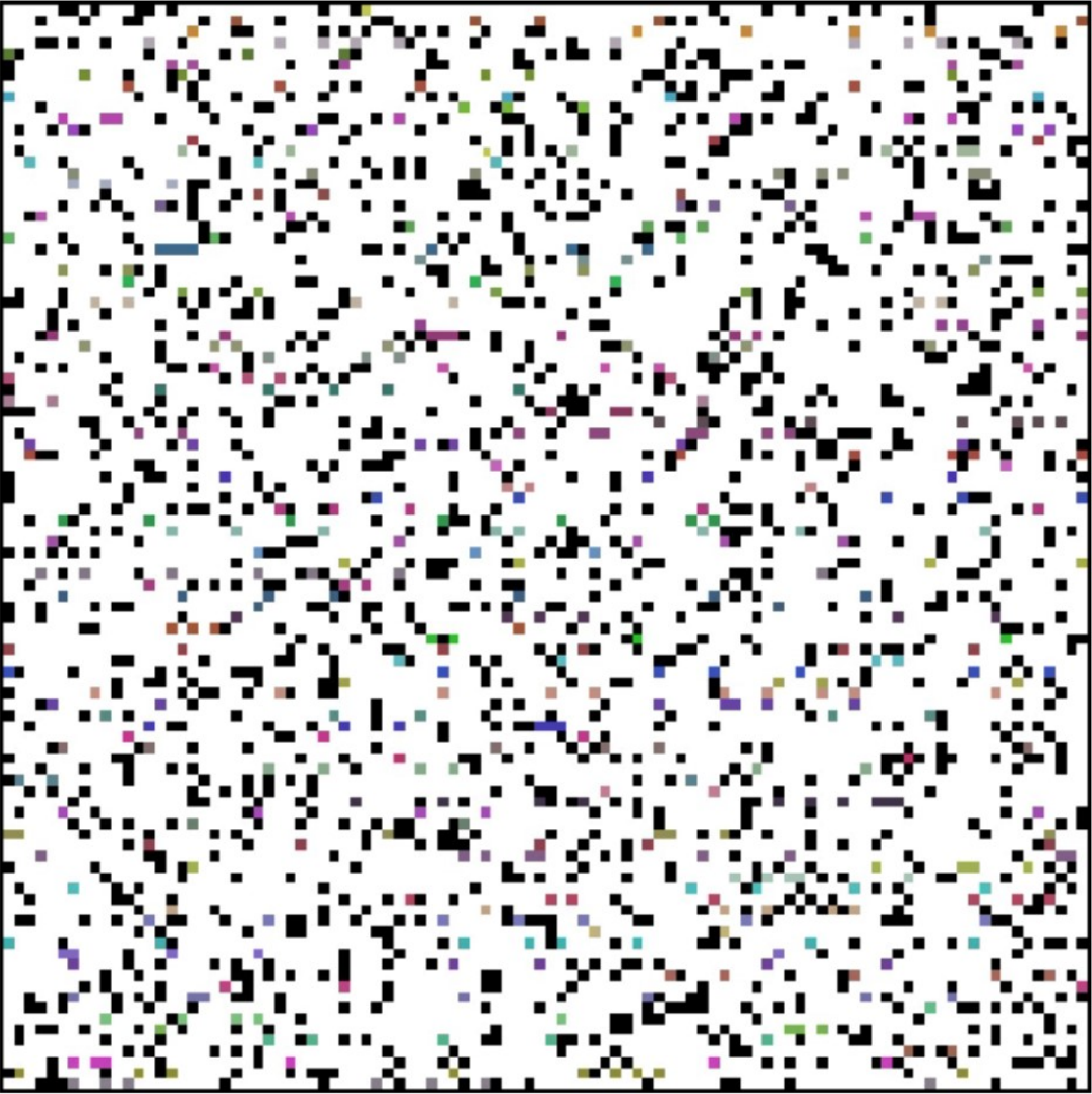}}
	\subfigure[Free map]{\includegraphics[width=0.3\columnwidth]{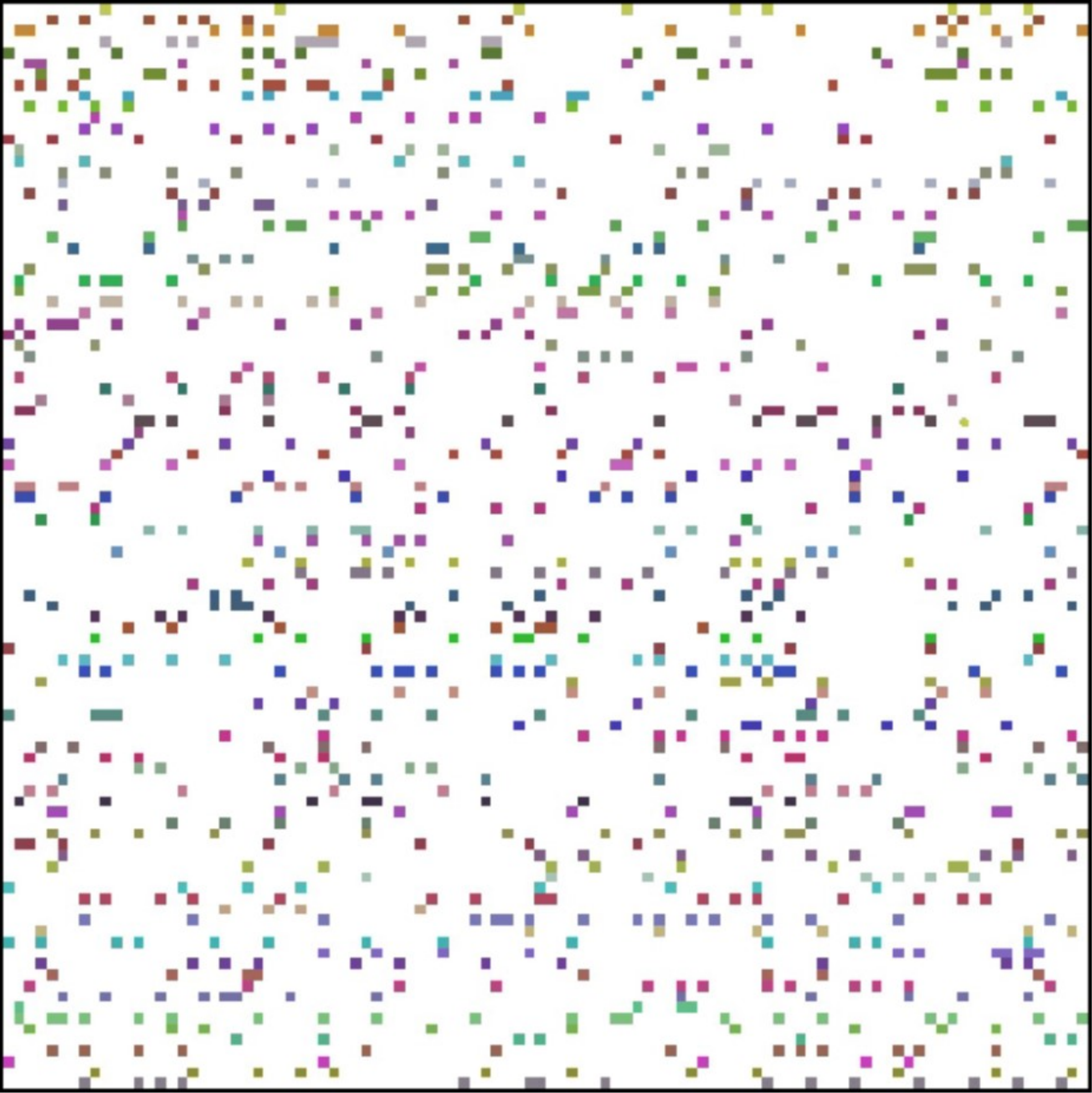}}
	\caption{Map examples. The black and colored nodes represent the static and dynamic obstacles respectively. Map parameters can be found in Section \ref{sec_map}.}
	\label{Fig_map}
	\vspace{-0.4cm}
\end{figure}

\subsection{Environments}\label{sec_map}
As shown in Figure \ref{Fig_map}, we consider three different environment maps to validate our approach, i.e., a regular map, a random map, and a free map. The first one imitates warehouse environments and contains both static and dynamic obstacles, where the static obstacles are regularly arranged and the dynamic ones move within the aisles. In the random map, we randomly set up a certain density of static obstacles and dynamic obstacles. In the free map, we only consider a certain density of dynamic obstacles. The default size of all the maps is $100\times100$. The static obstacle density in each map is set to $0.392$, $0.15$, and $0$, respectively, and the dynamic obstacle density is set to $0.03$, $0.05$, and $0.1$, respectively. Dynamic obstacles are modeled as un-controllable robots that are able to move one cell at each step in any direction. Their start/goal cells are randomly generated, and their desired trajectories are calculated through A* by considering the current position of any other obstacles. During training and testing, each dynamic obstacle continuously moves along its trajectory, and when motion conflict occurs, it will: 1) with a probability of 0.9, stay in its current cell until the next cell is available; 2) otherwise, reverse its direction and move back to its start cell.

\subsection{Training and Testing}
We train our model with one NVIDIA GTX 1080ti GPU in Python 3.6 with TensorFlow 1.4 \cite{tensorflow2015-whitepaper}. The learning rate is fixed to $3 \times 10^{-5}$, and the training optimizer is \texttt{RMSprop}. We use $\varepsilon$-greedy to balance exploration and exploitation. The initial value is set to be $\varepsilon=1$ and decreases to $0.1$ linearly when the total training steps reach $200,000$. In training, we randomly choose one of the three maps as the training map and configure the dynamic obstacles by following the settings in Section \ref{sec_map}. Then we randomly select two free cells as the start and goal cells. During the training, we end the current episode and start a new one if one of the following conditions is satisfied: 1) the number of training steps in the current episode reaches a maximum defined as $N_m=50+10\times N_e$; 2) the robot can not obtain any global guidance information in its local FOV; 3) the robot reaches its goal cell $c_{goal}$. In addition, after the robot completes $50$ episodes, the start and goal cells of all the dynamic obstacles are re-randomized.

Before learning starts, one robot explores the environment to fill up the replay buffer, which is comprised of a \texttt{Sumtree} structure to perform prioritized experience replay \cite{schaul2015prioritized}. Note that \texttt{Sumtree} is a binary tree, which computes the sum of the values of its children as the value of a parent node. In each episode, the robot samples a batch of transitions from the replay buffer with prioritized experience replay (PER) based on the calculated temporal difference error by the DDQN algorithm. During testing, all methods are executed on an Intel i7-8750H CPU.

\subsection{Performance Metrics}
The following metrics are used for performance evaluation:
\begin{itemize}[leftmargin=*]
\item{\it Moving Cost}:
\begin{equation}
\text{Moving Cost} = \frac{N_s}{||c_{goal}- c_{start}||_{L1}}
\end{equation}
where $N_s$ is the number of steps taken and $||c_{goal}- c_{start}||_{L1}$ is the Manhattan distance between the start cell and the goal cell. This metric is used to indicate the ratio of actual moving steps w.r.t the ideal number of moving steps without considering any obstacles.
\item{\it Detour Percentage}:
\begin{equation}
\text{Detour Percentage} = \frac{N_s-L_{A^*}(c_{start},c_{goal})}{L_{A^*}(c_{start},c_{goal})} \times 100\%
\end{equation}
where $L_{A^*}(c_{start},c_{goal})$ is the length of the shortest path between the start cell and the goal cell, which is calculated with A* algorithm by only considering the static obstacles. This metric indicates the percentage of detour w.r.t the shortest path length.
\item{\it Computing Time}: This measure corresponds to the average computing time at each step during the testing.
\end{itemize}

\begin{table*}[!t]
\renewcommand{\arraystretch}{1.2}
\caption{Single robot path planning results: Moving costs and detour percentages of different approaches}
\label{Tab_singlecompare}
\vspace{-0.2cm}
\centering
\begin{tabular}{c|cccc|cccc|ccc}
\hline
\hline
  &\multicolumn{4}{c|}{Moving Cost} & \multicolumn{4}{c|}{Detour Percentage} &\multicolumn{3}{c}{Computing Time (s)}  \\
\hline
 & Local & Global & Naive & G2RL & Local & Global & Naive & G2RL & Local & Global & G2RL \\
\hline
Regular-50  & 1.58(0.50) & 1.31(0.29) & 1.38(0.35) & $\textbf{1.18(0.16)}$ & 36.7(46)$\%$ & 23.7(27)$\%$ & 31.5(34)$\%$ & $\textbf{15.2(15)$\%$}$ & 0.004 & $\textbf{0.003}$ & 0.011 \\
Regular-100 & 1.57(0.35) & 1.23(0.21) & 1.42(0.31) & $\textbf{1.12(0.12)}$ & 36.3(34)$\%$ & 18.7(20)$\%$ & 39.5(30)$\%$ & $\textbf{10.7(12)$\%$}$ & 0.005 & $\textbf{0.004}$ & 0.012 \\
Regular-150 & 1.50(0.32) & 1.19(0.14) & 1.36(0.26) & $\textbf{1.09(0.08)}$ & 33.3(32)$\%$ & 16.0(14)$\%$ & 35.0(36)$\%$ & $\textbf{8.2(8)$\%$}$ & 0.007 & $\textbf{0.004}$ & 0.015  \\

Random-50  & 1.35(0.28) & 1.28(0.18) & 1.36(0.28) & $\textbf{1.21(0.13)}$ & 25.1(26)$\%$ & 21.1(17)$\%$ & 30.1(35)$\%$ & $\textbf{16.7(12)$\%$}$ & 0.005 & $\textbf{0.004}$ & 0.013  \\
Random-100 & 1.43(0.27) & 1.26(0.14) & 1.34(0.29) & $\textbf{1.15(0.10)}$ & 30.0(26)$\%$ & 20.5(14)$\%$ & 32.3(34)$\%$ & $\textbf{13.0(10)$\%$}$ & 0.006 & $\textbf{0.006}$ & 0.015 \\
Random-150 & 1.37(0.17) & 1.17(0.09) & 1.40(0.28) & $\textbf{1.11(0.08)}$ & 27.0(17)$\%$ & 14.5(9)$\%$ & 39.4(40)$\%$ & $\textbf{9.1(8)$\%$}$ & 0.010 & $\textbf{0.006}$ & 0.018 \\
Free-50  & 1.27(0.20) & 1.24(0.15) & 1.31(0.24) & $\textbf{1.14(0.09)}$ & 21.2(20)$\%$ & 19.4(15)$\%$ & 28.9(23)$\%$ & $\textbf{12.3(9)$\%$}$ & 0.008 & $\textbf{0.007}$ & 0.018  \\
Free-100 & 1.31(0.13) & 1.21(0.11) & 1.34(0.25) & $\textbf{1.11(0.07)}$ & 23.6(13)$\%$ & 17.3(11)$\%$ & 33.6(25)$\%$ & $\textbf{9.1(7)$\%$}$ & 0.015 & $\textbf{0.011}$ & 0.022  \\
Free-150 & 1.27(0.12) & 1.14(0.06) & 1.32(0.22) & $\textbf{1.07(0.05)}$ & 21.2(12)$\%$ & 12.3(6)$\%$ & 31.5(22)$\%$ & $\textbf{6.5(5)$\%$}$ & 0.017 & $\textbf{0.013}$ & 0.028  \\
\hline
\hline
\end{tabular}
\begin{tablenotes}
    \small
    \centering
    \item[*] Values are listed as ``mean (standard deviation)" across 100 instances. The lowest (best) values are highlighted.
\end{tablenotes}
\vspace{-0.2cm}
\end{table*}

\section{Results}
In this section, we present comparative results for both the single-robot and multi-robot path planning tasks. More details of our results can be found in the video \url{https://youtu.be/KbAp38QYU9o}.

\subsection{Single-Robot Path Planning Results}
We compare our approach with dynamic A* based methods with global re-planning and local re-planning strategies, and we call these two methods \texttt{Global Re-planning} and \texttt{Local Re-planning} respectively. For \texttt{Global Re-planning}, each time the robot encounters a conflict, an alternative path is searched for from the current cell to the goal cell by using the A* method, considering the current position of all the dynamic obstacles. For \texttt{Local Re-planning}, an alternative path is searched for from the current cell to the farthest cell within the robot's FOV. We also compare our reward function with a naive reward function, which strictly encourages the robot to follow the global guidance. Concretely, if the robot’s next location is on the global guidance, we give a constant positive reward $R(t)=0.01$. Otherwise, we give a large negative reward $R(t)=r_1+D_r\times r_4$, where $r_4=-0.03$ and $D_r$ is computed by calculating the distance between the robot’s next location with the nearest global guidance cell. We set a time-out for all the tests, within which time if the robot can not reach its goal, this test is defined as a failure case. In each test, the time-out value is set as the double of the Manhattan distance between the start cell and the goal cell of the robot. We train our model and the naive reward based model by using both the three environments shown in Figure \ref{Fig_map}, and then compare our testing results with \texttt{Global Re-planning} and \texttt{Local Re-planning} in the three environments separately. For each map, we separate the comparison into three groups with different Manhattan distances between the start cell and the goal cell, which are set to $50$, $100$, and $150$, respectively. For each group, $100$ pairs of start and goal locations are randomly selected and the mean value and its standard deviation are calculated. The desired trajectories of dynamic obstacles are consistent among the testing of each method for a fair comparison.

The comparison results in Table \ref{Tab_singlecompare} validate that: 1) Compared with \texttt{Local Re-planning} and \texttt{Global Re-planning}, our approach uses the smallest number of moving steps in all the cases. 2) Our approach has the smallest standard deviations in all the cases, which demonstrates our consistency across different settings.
3) Since under the naive reward function, the robot is encouraged to follow the global guidance strictly, its performance is worse than when utilizing our reward function. The naive reward introduces more detour steps and waiting time steps in the presence of motion conflicts, and may even cause deadlocks (which further lead to failure cases).
In contrast, our reward function incites convergence of the navigation tasks while simultaneously encouraging the robot to explore all the potential solutions to reach the goal cell with the minimum number of steps. Note that the \texttt{Moving Cost} and \texttt{Detour Percentage} are calculated by only considering the successful cases. The range of success rates for the naive reward based approach lies between $68\%-89\%$, whereas for ours it is consistently $100\%$. 4) Our computing time is slightly higher than \texttt{Local Re-planning} and \texttt{Global Re-planning}, but still remains within an acceptable interval. Our maximum computing time is about $28ms$, which means that our RL planner achieves an update frequency of more than $35Hz$ on the given CPU platform, fulfilling the real-time requirements of most application scenarios.

\begin{table}[h]
\renewcommand{\arraystretch}{1.2}
\caption{Ablation study on different robot FOV sizes}
\label{Tab_FOV}
\vspace{-0.2cm}
\centering
\begin{tabular}{c|ccccc}
\hline
\hline
$H_l\times W_l$ & 7$\times$7 & 9$\times$9 & 11$\times$11 & 13$\times$13 & 15$\times$15\\
\hline
Random-50   & 1.49 & 1.35 & 1.32 & 1.24 & 1.21 \\
Random-100   & 1.33 & 1.25 & 1.23 & 1.17 & 1.15 \\
Random-150 & 1.27 & 1.21 & 1.18 & 1.14 & 1.13\\
\hline
\hline
\end{tabular}
\begin{tablenotes}
    \small
    \centering
    \item[*] Values show the mean of the Moving Cost index.
\end{tablenotes}
\vspace{-0.1cm}
\end{table}

\begin{table}[h]
\renewcommand{\arraystretch}{1.2}
\caption{Ablation study on different input sequence lengths}
\label{Tab_seq}
\vspace{-0.2cm}
\centering
\begin{tabular}{c|cccc}
\hline
\hline
$N_t$ & 1 & 2 & 3 & 4\\
\hline
Random-50 & 1.42 & 1.28 & 1.22 &1.21\\
Random-100 & 1.48 & 1.34 & 1.21 & 1.15\\
Random-150 & 1.56 & 1.31 & 1.19 & 1.13\\
\hline
\hline
\end{tabular}
\begin{tablenotes}
    \small
    \centering
    \item[*] Values show the mean of the Moving Cost index.
\end{tablenotes}
\vspace{-0.1cm}
\end{table}

We further test the effect of different robot FOV size and input sequence length $N_t$ on the performance. The random map in Figure \ref{Fig_map} is used, and the start and goal cells are generated with fixed Manhattan distances of $50$, $100$, and $150$. For each value, $100$ pairs of start and goal cells are tested. We first choose different values of the FOV size of $H_l\times W_l$ for comparison. The results in Table \ref{Tab_FOV} show that: 1) As the FOV size increases, the robot reaches the goal cell in less average steps. 2) The performance improvement is not significant when the FOV size is large than $13\times 13$. Since a smaller FOV size implies a smaller learning model, a FOV size of $15 \times 15$ is large enough for our cases to balance the performance and computation cost. We then compare different values of the input sequence length $N_t$ for comparison. For ease of implementation, we keep the same network input size in all the cases and use empty observation images for the cases with $N_t<4$. Note that if $N_t=1$, the robot loses all the temporal information of the dynamic obstacles and only considers the current observation. The results in Table \ref{Tab_seq} show that: 1) Introducing the historical local observation information improves the system performance in all the cases significantly. 2) When $N_t>3$, increasing $N_t$ only marginally improves the performance. Since a smaller $N_t$ implies a smaller learning model, $N_t=4$ is large enough for our cases to balance the performance and computation cost.

\begin{table}[h]
\renewcommand{\arraystretch}{1.2}
\caption{Experimental results in unseen environments}
\label{Tab_unseen}
\vspace{-0.2cm}
\centering
\begin{tabular}{c|ccc}
\hline
\hline
 & Local & Global & G2RL\\
\hline
Moving Cost & 1.46(0.19) & 1.21(0.08) & $\textbf{1.12(0.06)}$\\
Detour Percentage & 42.8(19)$\%$ & 20.5(8)$\%$ & $\textbf{9.6(6)$\%$}$\\
\hline
\hline
\end{tabular}
\vspace{-0.1cm}
\end{table}

\begin{table*}[!t]
\renewcommand{\arraystretch}{1.2}
\caption{Multi-robot path planning results: success rate and computational time of different approaches}
\label{Tab_multi}
\vspace{-0.2cm}
\centering
\begin{tabular}{c|cccccc|ccc}
\hline
\hline
{} & \multicolumn{6}{c|}{Success Rate} & \multicolumn{3}{c}{Computing Time (ms)} \\
\hline
 & ECBS & HCA* & Global Re-planning & Discrete-ORCA & PRIMAL & G2RL & Discrete-ORCA & PRIMAL & G2RL \\
\hline
Regular & $\textbf{100$\%$}$ & $\textbf{100$\%$}$ & 95.7$\%$ & 88.7$\%$ & 92.3$\%$ & 99.7$\%$ & \textbf{2.064(0.107)} & 5.892(0.104) & 6.367(0.337)\\

Random & $\textbf{100$\%$}$ & $\textbf{100$\%$}$ & 98.2$\%$ & 55.0$\%$ & 80.6$\%$ & 99.7$\%$ & \textbf{1.233(0.044)} & 6.620(0.280) & 6.206(0.260)\\

Free & $\textbf{100$\%$}$ & $\textbf{100$\%$}$ & 98.8$\%$ & 99.5$\%$ & 75.7$\%$ & 99.8$\%$ & \textbf{1.480(0.050)} & 7.319(0.300) & 6.246(0.277)\\

\hline
\hline
\end{tabular}
\end{table*}

\begin{figure*}[!t]
	\centering
    \subfigure[Regular map with 32 robots]{\includegraphics[width=0.67\columnwidth]{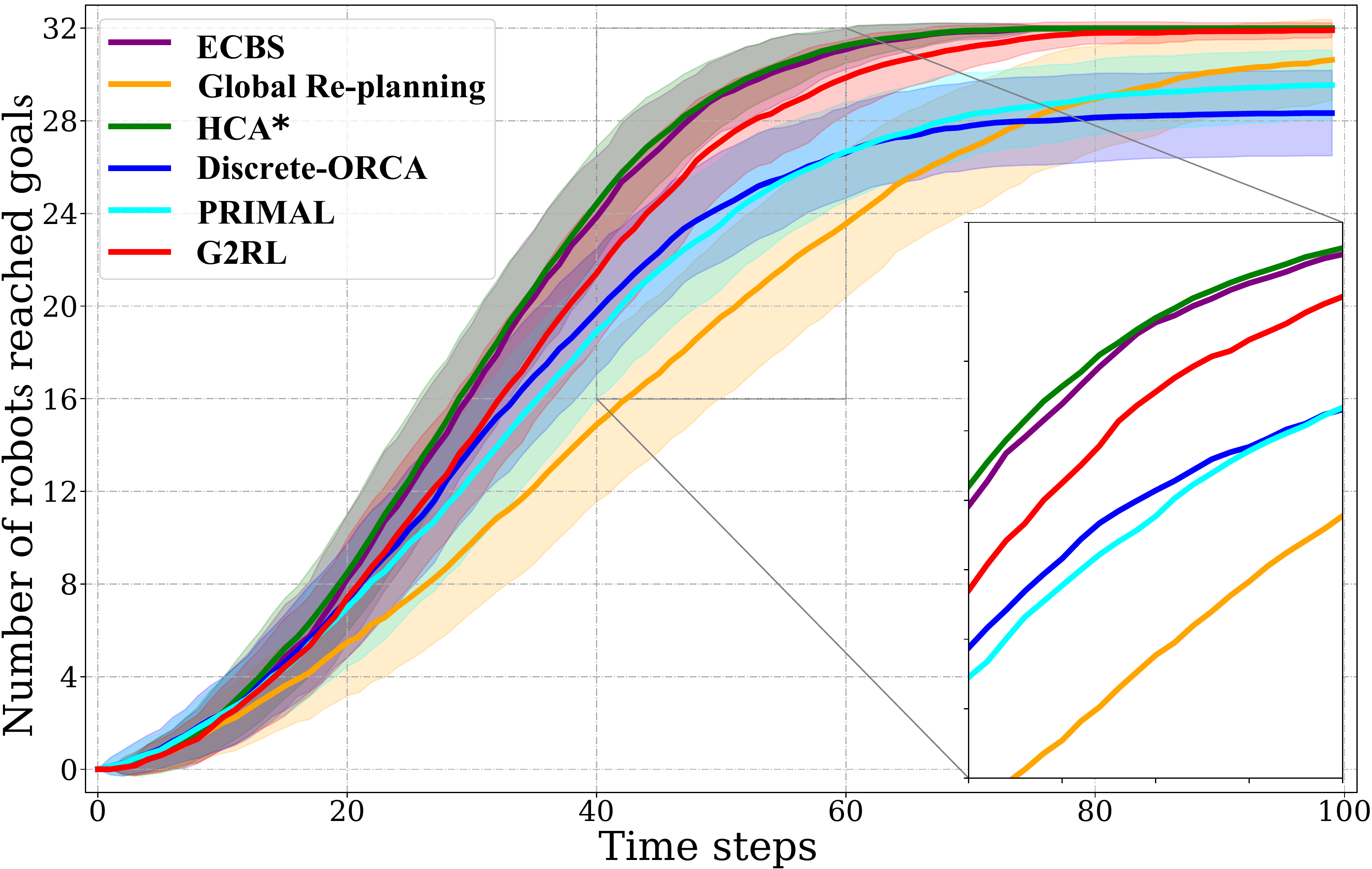}}
    \subfigure[Random map with 64 robots]{\includegraphics[width=0.67\columnwidth]{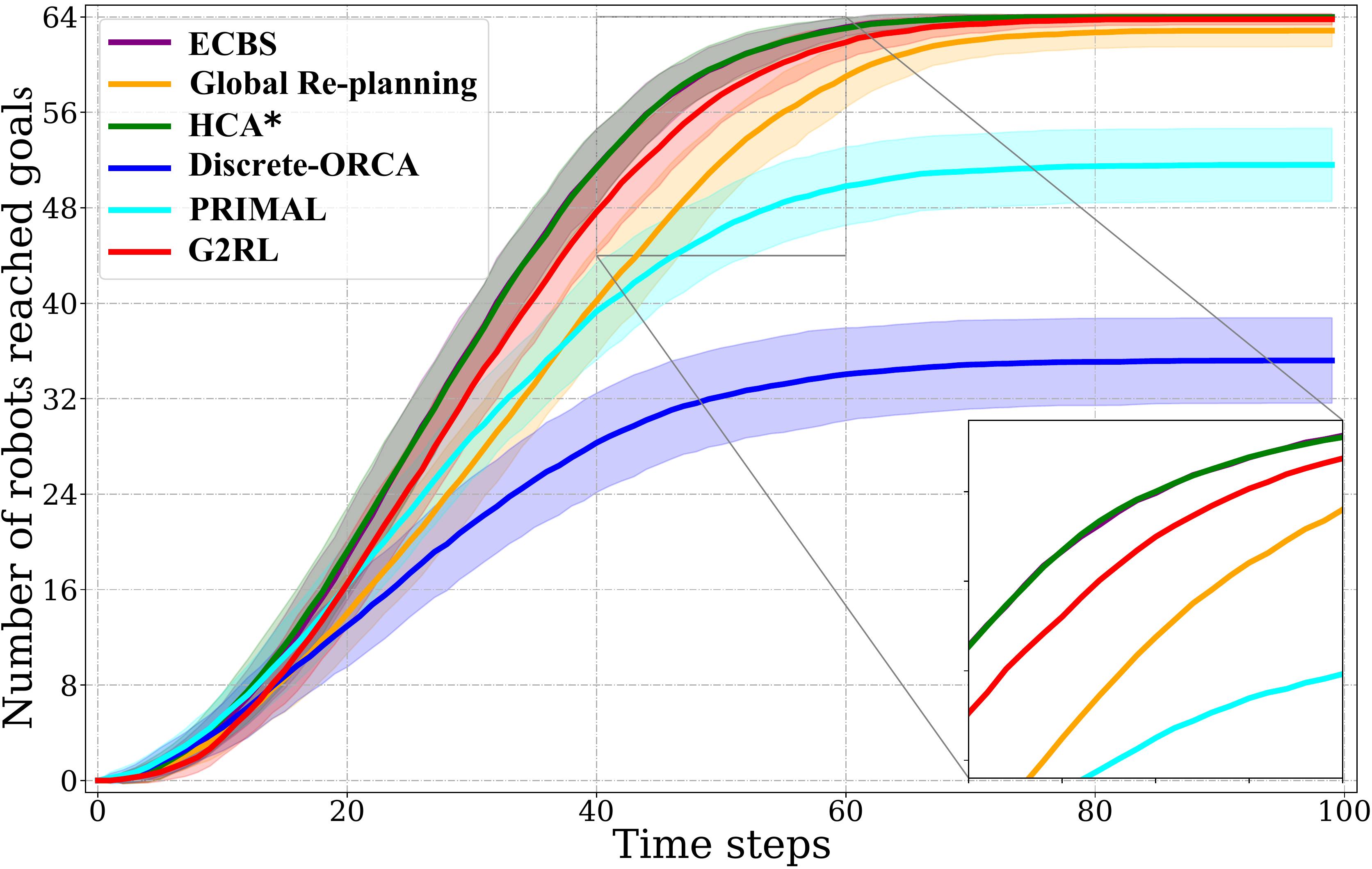}}
    \subfigure[Free map with 128 robots]{\includegraphics[width=0.67\columnwidth]{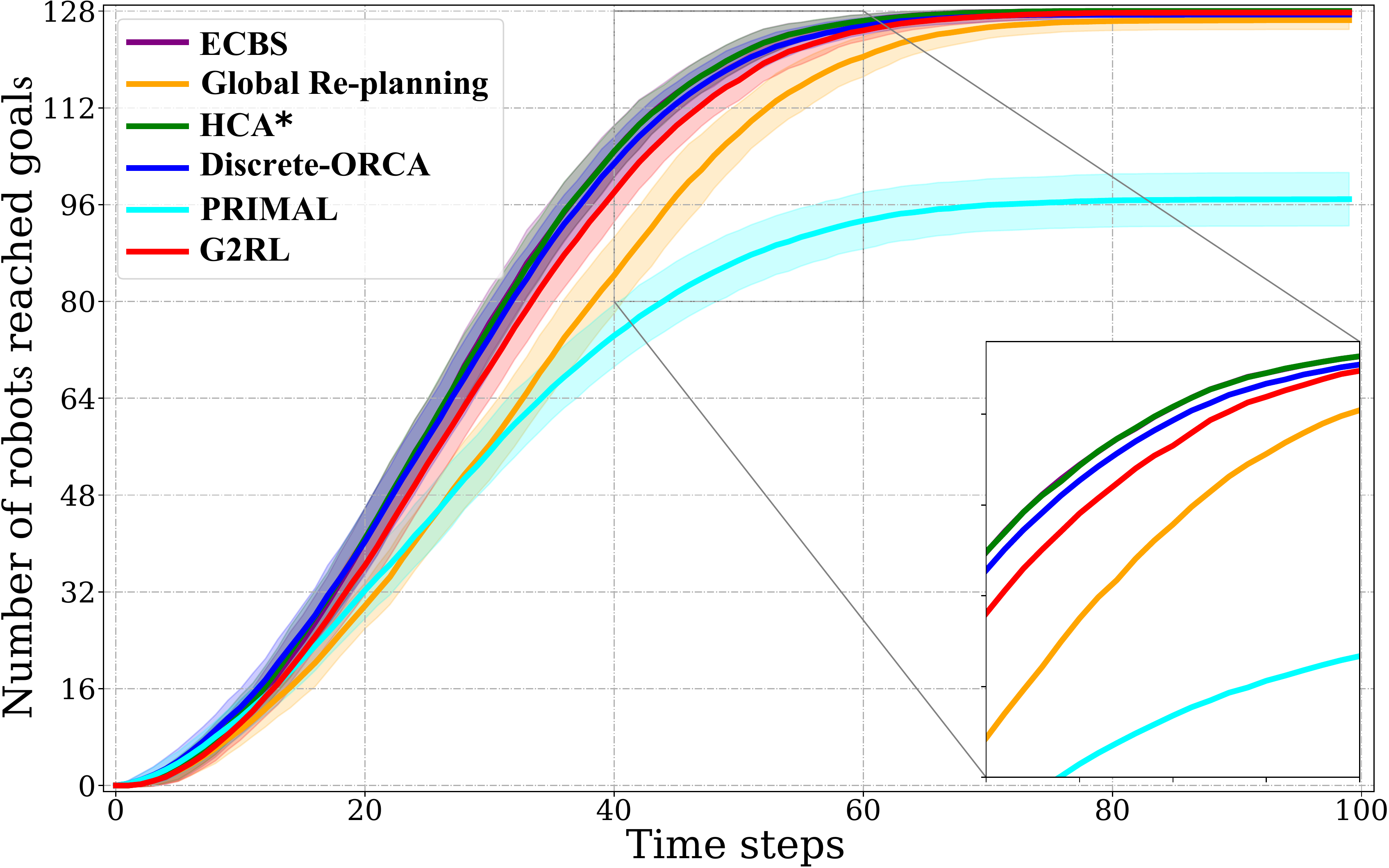}}
    \subfigure[Regular map with 32 robots]{\includegraphics[width=0.67\columnwidth]{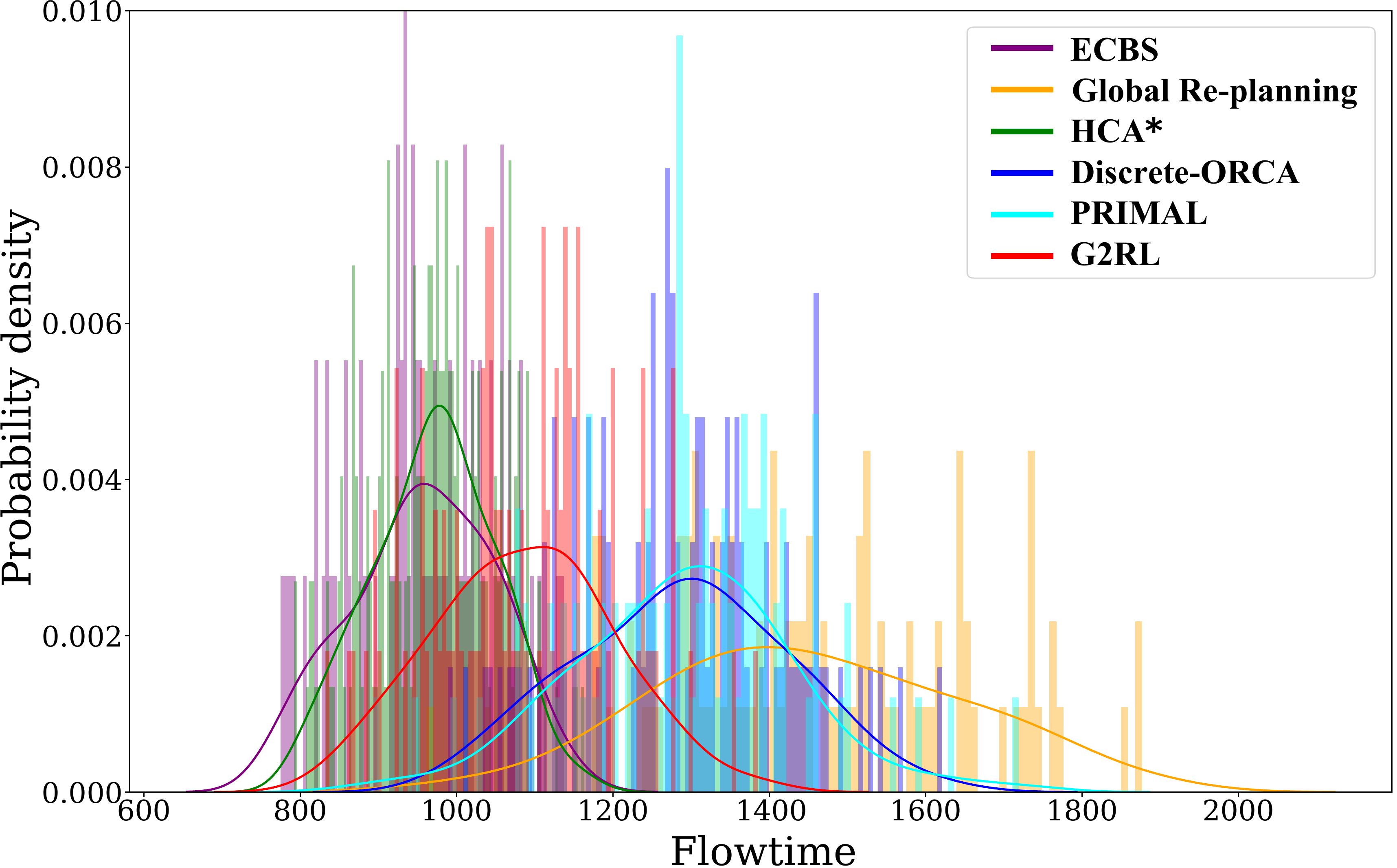}}
    \subfigure[Random map with 64 robots]{\includegraphics[width=0.67\columnwidth]{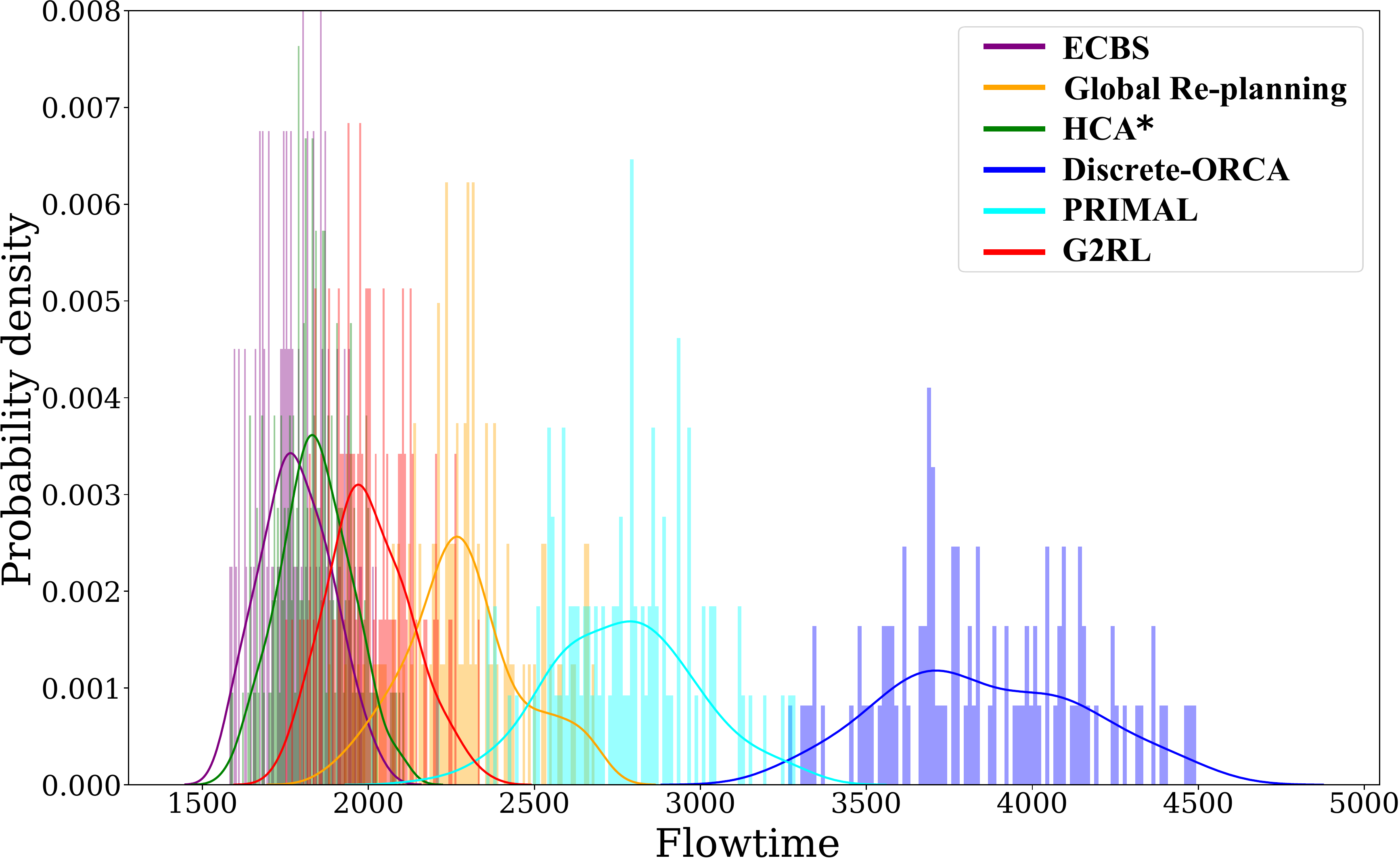}}
    \subfigure[Free map with 128 robots]{\includegraphics[width=0.67\columnwidth]{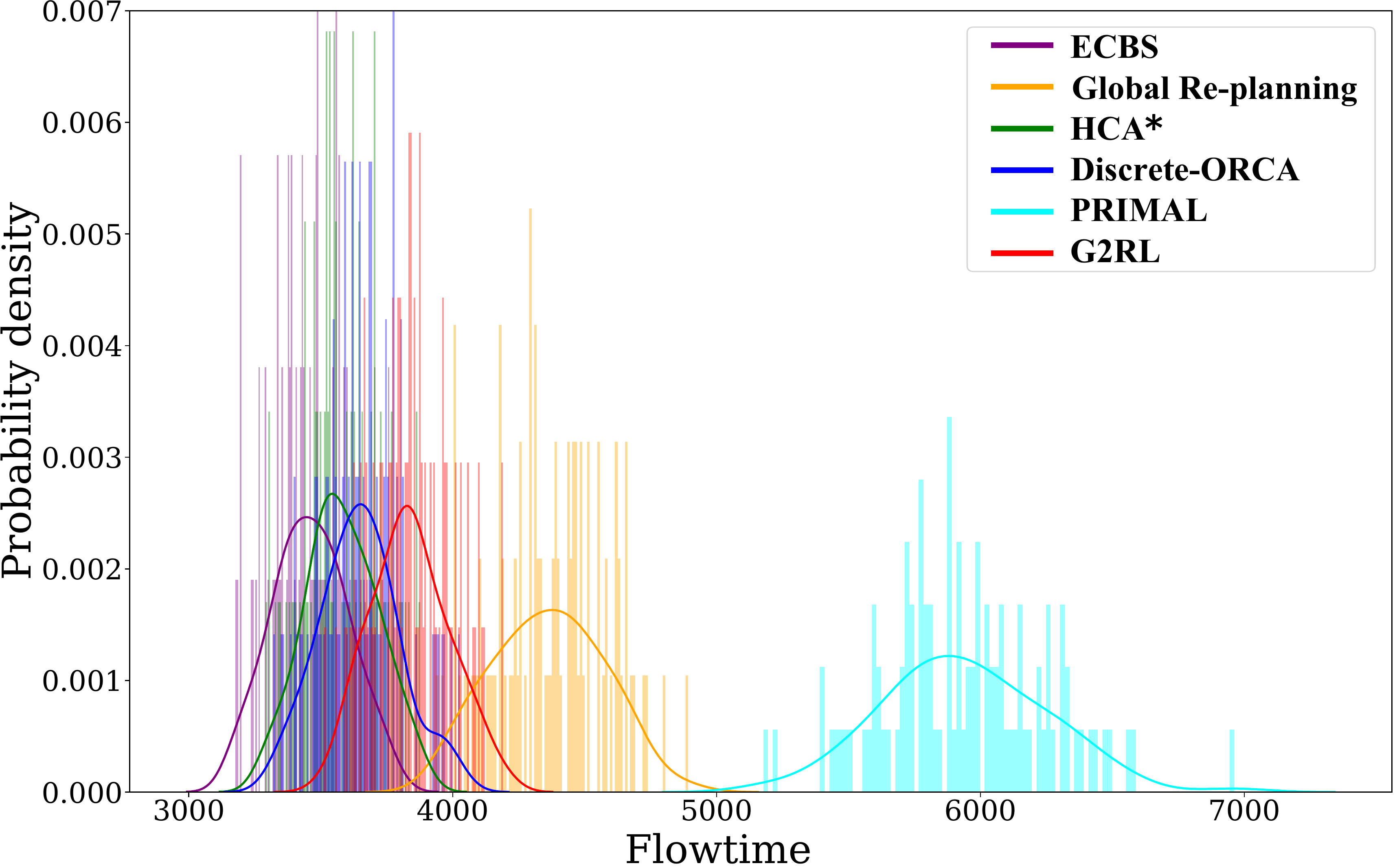}}
	\caption{Multi-robot path planning results. The upper row shows the number of reached robots at different steps of different approaches in three testing maps. The solid lines show the average number across 100 tests, and the shadow areas represent the standard deviations. The lower row plots the corresponding histograms of flowtime values. The flowtime of all the failure cases is set to the maximum time step $100$.}
	\label{fig:Fig_multi}
\end{figure*}

Next, we test our approach in an unseen environment to validate its generalizability. The environment is an enlarged version of the random map in Figure \ref{Fig_map} with a size $200 \times 200$. The densities of static and dynamic obstacles are set to $0.15$ and $0.05$, respectively. We randomly generate 100 pairs of start and goal cells with a constant Manhattan distance of $200$. Note that we directly use the model trained in the small maps for testing. The results in Table \ref{Tab_singlecompare} and Table \ref{Tab_unseen} show that our approach performs consistently well, under different environments, which validates the scalability and generalizability of our approach.

\subsection{Comparison with Multi-Robot Path Planning Methods}
We apply our method to the problem of multi-robot path planning, and compare it to five state-of-the-art benchmarks: \textit{(i)} a decoupled dynamic A* based method, \texttt{Global Re-planning}, \textit{(ii)} a decoupled path planning method, \texttt{HCA}$^*$\cite{Silver2005Cooperative}, \textit{(iii)} a centralized optimal path planning method, Conflict-Based Search (\texttt{CBS})\cite{barer2014suboptimal}, \textit{(iv)} a velocity-based method \texttt{ORCA}\cite{Jur2011Reciprocal} and, \textit{(v)}, the RL based approach \texttt{PRIMAL}\cite{sartoretti2019primal}. For \texttt{HCA}$^*$, the priorities of the robots are randomly chosen.
For \texttt{CBS}, since computing an optimal solution when the robot number is larger than $50$ is often intractable, we use its sub-optimal version \texttt{ECBS} \cite{barer2014suboptimal} with a sub-optimality bound set to $1.03$. For \texttt{ORCA}, we calculate the next position of the robot by only considering the angle of the velocity output and thus transform the continuous-space \texttt{ORCA} into a discrete version \texttt{Discrete-ORCA}. It should be noted that neither our approach nor \texttt{PRIMAL} has been trained in the testing maps. For \texttt{PRIMAL}, we directly use the trained model provided online by the authors \footnote{\url{https://github.com/gsartoretti/distributedRL\_MAPF}}. In our approach, we directly use the model trained in a single robot case.

Comparisons are performed in three different maps with size $40\times 40$ and varying numbers of robots, which are smaller versions of the maps in Figure \ref{Fig_map}. The static obstacle densities are set to $0.45$, $0.15$, and $0$ in the regular, random, and free maps, respectively, and the robot numbers are set to $32$, $64$, and $128$. In each map, we generate $100$ random configurations of robots with different start and goal cells. To ensure the solvability of the problem, once each robot arrives at its goal cell, we remove the robot from the environment to avoid conflicts. We set a time-out of $100$ time-steps for all the tests, within which time, if any robot can not reach its goal, this test is defined as a failure case.
In Figures \ref{fig:Fig_multi} (a)-(c), we compare the number of robots that have reached their goals as a function of time. In Figures \ref{fig:Fig_multi} (d)-(f), we compare the flowtime of the different approaches, which is defined as the sum of traversal time steps across all the robots involved in the instance. In Table \ref{Tab_multi}, we compare their success rates and computing times. Since in \texttt{Global Re-planning}, \texttt{ECBS} and \texttt{HCA}$^*$, the robot action is not generated online at each step, we only compare the computing time of \texttt{Discrete-ORCA}, \texttt{PRIMAL} and our approach. The computing time is normalized by the average flowtime.

These results show that: 1) Our approach maintains consistent performance across different environments, outperforming \texttt{Global Re-planning} and \texttt{PRIMAL}, also outperforming \texttt{Discrete-ORCA} in most cases (\texttt{Discrete-ORCA} can not handle the crowded static obstacles, so it is only effective in the free map). \texttt{ECBS} and \texttt{HCA}$^*$ achieve the best performance overall, since they are both centralized approaches that have access to all robots' trajectory information. In \texttt{HCA}$^*$, the trajectory information of all robots with higher priorities is shared and utilized in the planning of the robots with lower priorities. Our approach is in stark contrast to these approaches, since it is fully distributed and non-communicative (i.e., requiring no trajectory information of other robots). 2) Our success rate is similar to that of \texttt{ECBS} and \texttt{HCA}$^*$, and higher than \texttt{Global Re-planning}, \texttt{Discrete-ORCA} and \texttt{PRIMAL}. The results show that our approach outperforms all the distributed methods, which validates its robustness, scalability, and generalizability. We note that in contrast to \texttt{PRIMAL}, which uses a general `direction vector'\cite{sartoretti2019primal}, we utilize the global guidance instead. The global guidance provides dense global information which considers all the static obstacles. Thus, our method is able to overcome many of the scenarios (e.g., deadlocks) where \texttt{PRIMAL} gets stuck.

\section{Conclusion}
In this paper, we introduced G2RL, a hierarchical path planning approach that enables end-to-end learning with a fixed-sized model in arbitrarily large environments. We provided an application of our approach to distributed uncoupled multi-robot path planning that is naturally scaled to an arbitrary number of robots. Experimental results validated the robustness, scalability, and generalizability of this path planning approach. Notably, we demonstrated that its application to multi-robot path planning outperforms existing distributed methods and that it performs similarly to state-of-the-art centralized approaches that assume global dynamic knowledge. In future work, we plan to extend our approach to cooperative multi-robot path planning.

\bibliographystyle{IEEEtran}
\bibliography{references_final}

\ifCLASSOPTIONcaptionsoff
  \newpage
\fi

\end{document}